\documentclass[lettersize,journal]{IEEEtran}
\usepackage{amsmath,amsfonts,amssymb}
\usepackage{algorithmic}
\usepackage{algorithm}
\usepackage{array}
\usepackage{subfig}
\usepackage{textcomp}
\usepackage{stfloats}
\usepackage{url}
\usepackage{verbatim}
\usepackage[dvipdfmx]{graphicx}
\usepackage{cite}
\usepackage{color}
\usepackage{xcolor}
\usepackage{subfiles}
\usepackage{comment}
\usepackage{bm}
\usepackage{cases}

\newtheorem{theorem}{Theorem}

\newtheorem{lemma}{Lemma}
\newtheorem{remark}{Remark}

\newtheorem{proposition}{Proposition}
\newtheorem{definition}{Definition}

\DeclareMathOperator*{\argmin}{arg\,min}

\newcommand{\mbr}{\mathbb{R}}
\newcommand{\diag}{\mathrm{diag}}

\newcommand{\mk}{\mkern -1mu}

\graphicspath{{figure/}}

\newcommand{\SO}{\mathrm{SO}(3)}
\DeclareMathOperator{\rank}{rank}
\DeclareMathOperator{\cond}{cond}
\DeclareMathOperator{\Null}{Null}
\begin{document}

\title{A Two-Wheeled Quadrotor with All-Clockwise Tilted Rotors for Level-Attitude Surface Locomotnion}

\author{Takumi Ito$^{1}$, 
Satoshi Nakano$^{2}$% <-this % stops a space
\thanks{*This work was supported by JSPS KAKENHI Grant Numbers JP26K24610, JP23K13352.}% <-this % stops a space
\thanks{$^{1}$ National Institute of Advanced Industrial Science and Technology (AIST), 
    {\tt\small  takumi.ito@aist.go.jp}}%
\thanks{$^{2}$ Nagoya Institute of Technology, 
    {\tt\small  nakano@nitech.ac.jp}}%
}
\markboth{Journal of \LaTeX\ Class Files,~Vol.~14, No.~8, August~2021}%
{Shell \MakeLowercase{\textit{et al.}}: A Sample Article Using IEEEtran.cls for IEEE Journals}

\IEEEpubid{0000--0000/00\$00.00~\copyright~2021 IEEE}

\maketitle

\begin{abstract}
Conventional wheel-equipped quadrotor UAVs must tilt their bodies to generate lateral forces, limiting their ability to maintain a desired attitude during lateral motion. This paper proposes a wheeled quadrotor with all-clockwise tilted rotors that uses surface contact as a constraint to generate propulsion along the contact surface without tilting the vehicle. We analyze feasible forces and system controllability under surface contact and optimize the vehicle design considering different rotor-spin configurations. The optimization results show that a configuration in which all four rotors spin clockwise can generate a larger tangential force than the conventional alternating-spin configuration. This approach enables both aerial and contact-based locomotion without additional rotor-tilting actuators or mechanisms. Simulation results demonstrate both hovering in free space and horizontal locomotion on a floor while maintaining a level attitude.
\end{abstract}

% \begin{IEEEkeywords}
% Article submission, IEEE, IEEEtran, journal, \LaTeX, paper, template, typesetting.
% \end{IEEEkeywords}

%%============================================================================%%
\section{Introduction}

Unmanned aerial vehicles (UAVs) can hover and operate in locations that are difficult or hazardous for humans. They are useful for tasks such as infrastructure inspection \cite{Jordan2018survey}, disaster response \cite{yucesoy2025disaster}, and remote monitoring \cite{ahmad2025monitoring}. Most conventional UAVs have all rotors arranged on a common plane and oriented upward. Although this configuration efficiently generates upward force against gravity, it cannot directly generate lateral forces while maintaining a level attitude. Therefore, lateral motion generally requires the UAV to tilt its body. This limits tasks that require a specific body orientation, such as inspection while keeping a camera level. Another challenge arises when UAVs interact with their surroundings. Contact with walls or other surfaces can disturb the vehicle and make stable flight difficult. Thus, conventional UAVs face two main challenges: underactuated force generation and difficulty in maintaining stable flight during contact.

Several studies have addressed the latter challenge by incorporating protective structures into UAVs for safe physical interaction. For example, shells surrounding whole body \cite{Kornatowski2020morphing} or the propellers \cite{Salaan2019passiveshell}. Similarly, wheels have been incorporated into UAVs to enable locomotion while maintaining contact with the environment \cite{tanaka2025wheeldrone, zhang2023wheel}. Although these approaches improve the safety and robustness of contact-based operations, the fundamental limitation in lateral force generation remains.

Fully actuated UAVs have been developed to overcome this limitation. For example, by fixing the rotors at tilted angles, a hexacopter can generate forces in arbitrary directions \cite{flores2023hexa, tadokoro2018hexa}. However, this approach generally requires at least six independent control inputs to independently control the six-dimensional wrench, consisting of three forces and three moments. For quadrotors, actively tilting the rotors can provide full actuation with only four rotors by changing the thrust directions during flight \cite{li2026tiltable, ryll2012modeling}. However, this approach requires additional actuators and mechanical mechanisms for rotor tilting, resulting in increased mechanical complexity and weight.

\IEEEpubidadjcol
In this paper, we propose a novel wheeled quadrotor UAV that can hover in free space and move along a surface while maintaining contact with it. Two key ideas enable this capability. First, the proposed system has tilted rotors and uses surface contact as a constraint to overcome the underactuated force-generation problem. This enables the vehicle to generate propulsion along the contact surface without tilting its body. Second, we consider different rotor-spin configurations rather than restricting the design to the conventional alternating-spin configuration. Here, a rotor spinning clockwise is denoted as a \textit{P}-rotor, while a rotor spinning counterclockwise is denoted as an \textit{N}-rotor. Accordingly, the conventional alternating-spin configuration is denoted as \textit{PNPN}. Although alternating spin directions are commonly used in quadrotors, previous work has shown that a quadrotor with all four rotors spinning in the same direction, denoted as \textit{PPPP (4P)}, can also achieve hovering flight \cite{Ito19052023}. This motivates us to consider the 4P configuration for contact-based locomotion.

Based on a control-theoretic analysis of force and moment generation under surface contact, we formulate an optimization problem for both the PNPN and 4P configurations. The optimization results are compared in terms of the maximum tangential force that can be generated while maintaining contact with the surface. The results show that the 4P configuration can generate a larger tangential force than the PNPN configuration. This demonstrates that combining the 4P configuration with the contact constraint provides a simple way to achieve both aerial and contact-based locomotion without additional rotor-tilting actuators.

The main contributions of this work are summarized as follows:
1) A theoretical analysis of feasible forces and system controllability under contact constraints. The analysis considers general contact plane and is not limited to floors or walls.
2) Proposing an two-wheeled UAV with the 4P configuration and optimizing it for locomotion on a horizontal surface, considering both tangential force generation and system controllability.
3) Simulation validation comparing the proposed 4P configuration with a standard wheeled quadrotor during surface locomotion.

\section{Modeling}%===========================

\subsection{Coordinate Frames and Kinematics}

\label{sec:dynamics}
% \subsection{記号の定義}
% \label{subsec:definition}
\begin{figure}[t]
    \centering
    \includegraphics[width=0.5\linewidth]{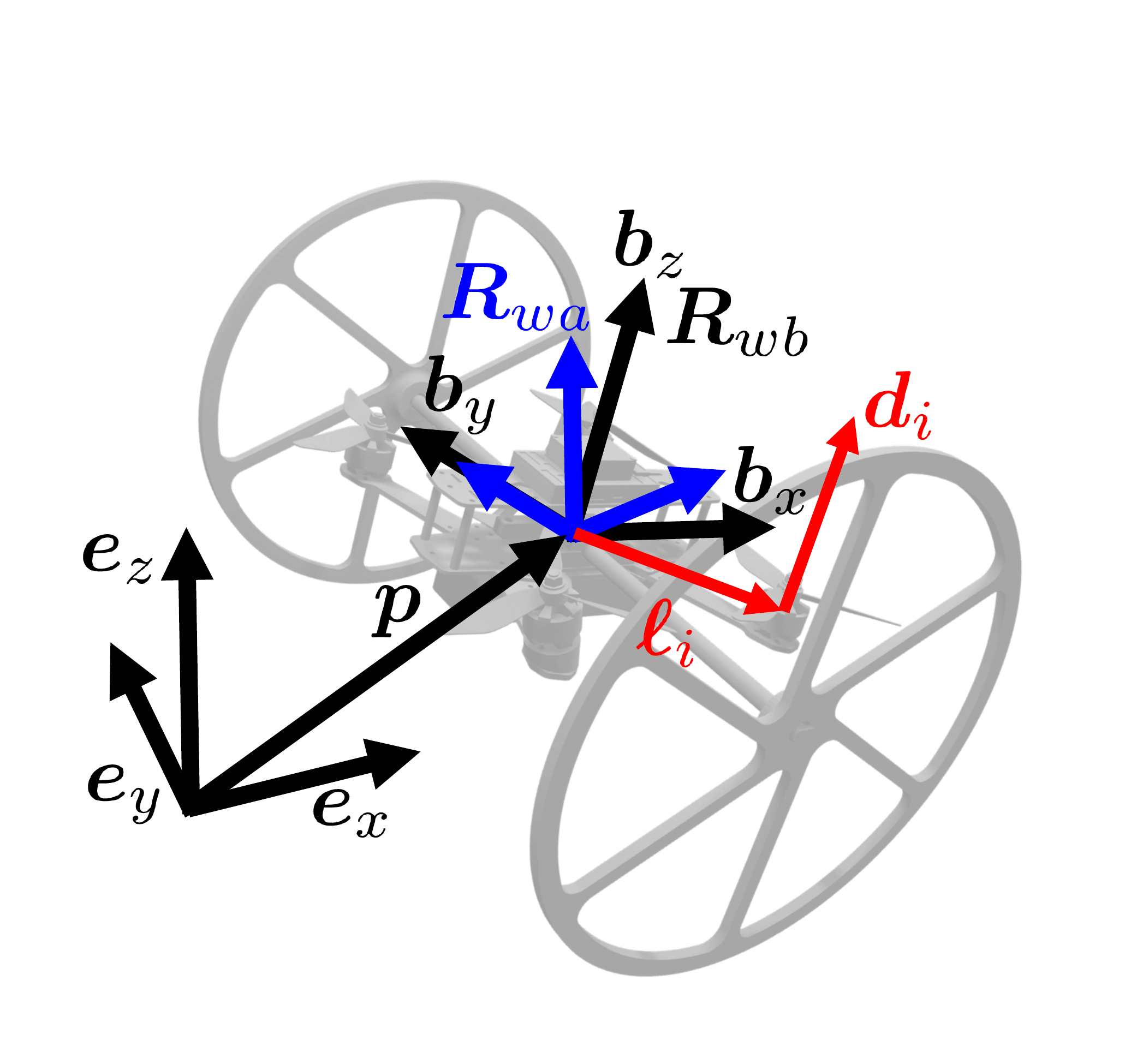}
    \caption{Coordinate frame of two-wheeled drone}
    \label{fig:two_wheel_drone_model}
\end{figure}

Let $\Sigma_w=(\bm e_x,\bm e_y,\bm e_z)$ denote the inertial frame with the standard basis of $\mbr^3$, and let $\Sigma_b=(\bm b_x,\bm b_y,\bm b_z)$ denote the body-fixed frame as shown in Fig.~\ref{fig:two_wheel_drone_model}.
The vectors $\bm b_x$ and $\bm b_y$ lie on the plane passing through the
rotor centers, with $\bm b_y$ aligned with the wheel axle; $\bm b_z$ is
normal to this plane and points in the nominal thrust direction.
The origin of $\Sigma_b$ is assumed to coincide with both the center of mass
and the axle center.
Let $m\in\mbr_{>0}$ be the mass, $\bm J=\diag(J_x,J_y,J_z)\in\mbr^{3\times3}$
the inertia matrix, $r_w\in\mbr_{>0}$ the wheel radius, and
$\ell_w\in\mbr_{>0}$ the distance from the body center to each wheel center.

Let the position and attitude of $\Sigma_b$ with respect to $\Sigma_w$ be
$\bm p\in\mbr^3$ and $\bm R=[\bm b_x~\bm b_y~\bm b_z]\in\SO$, 
and the body-frame linear and angular velocities be
$\bm v\in\mbr^3$ and $\bm\Omega\in\mbr^3$.
Then, the kinematics are
\begin{align}
\label{eq:kinematics}
    \dot{\bm p} = \bm R\bm v,
    \qquad
    \dot{\bm R} = \bm R\hat{\bm\Omega},
\end{align}
where $\hat{\cdot}:\mbr^3\to\mathfrak{so}(3)$ is the skew-symmetric operator
satisfying $\hat{\bm a}\bm b=\bm a\times\bm b$.

\subsection{Rotor Wrench Map}
\label{subsec:wrench_map}

The $i$-th of the $n_r$ rotors generates a thrust $f_i\in\mbr_{\ge0}$ with a proportional counter-torque \cite{lee2010},
\begin{align}
    \tau_i=\kappa_i f_i,
    \qquad
    \kappa_i=\pm\kappa,~\kappa\in\mbr_{>0},
    \label{eq:counter_torque}
\end{align}
where the sign of $\kappa_i$ is determined by the spin direction of the rotor: following \cite{Ito19052023}, we call a clockwise rotor \emph{positive} (P) and a counter-clockwise rotor \emph{negative} (N). 
Collecting the thrusts into the input vector $\bm u:=[f_1~\cdots~f_{n_r}]^\top$, restricted to the box $\mathcal U:=[0,f_{\max}]^{n_r}$.
The thrust and counter-torque are along its axis $\bm d_i\in\mathbb{S}^2$ at position $\bm\ell_i\in\mbr^3$, both expressed in $\Sigma_b$, then the total body-frame wrench is expressed as a linear map
\begin{align}
    \begin{bmatrix}
        \bm f\\ \bm\tau
    \end{bmatrix}
    =
    \sum_{i=1}^{n_r}
    \begin{bmatrix}
        \bm d_i\\
        \kappa_i\bm d_i+\bm\ell_i\times\bm d_i
    \end{bmatrix} f_i
    =\bm M\bm u,
    \label{eq:wrench_map}
\end{align}
where $\bm M\in\mbr^{6\times n_r}$ is the \emph{control effectiveness matrix}, partitioned as
\begin{align}
    \bm M=
    \begin{bmatrix}
        \bm M_f\\ \bm M_\tau
    \end{bmatrix},
    \quad
    \bm M_f=
    \begin{bmatrix}
        \bm M_{f,x}^\top\\ \bm M_{f,y}^\top\\ \bm M_{f,z}^\top
    \end{bmatrix},
    \quad
    \bm M_\tau=
    \begin{bmatrix}
        \bm M_{\tau,x}^\top\\ \bm M_{\tau,y}^\top\\ \bm M_{\tau,z}^\top
    \end{bmatrix}.
\end{align}
The wheels are passive and generate no driving torque; tangential propulsion during contact locomotion is therefore produced only by rotor thrust.
%, while the wheels impose the rolling constraints introduced below.

\begin{figure}[t]
    \centering
    \includegraphics[width=\linewidth]{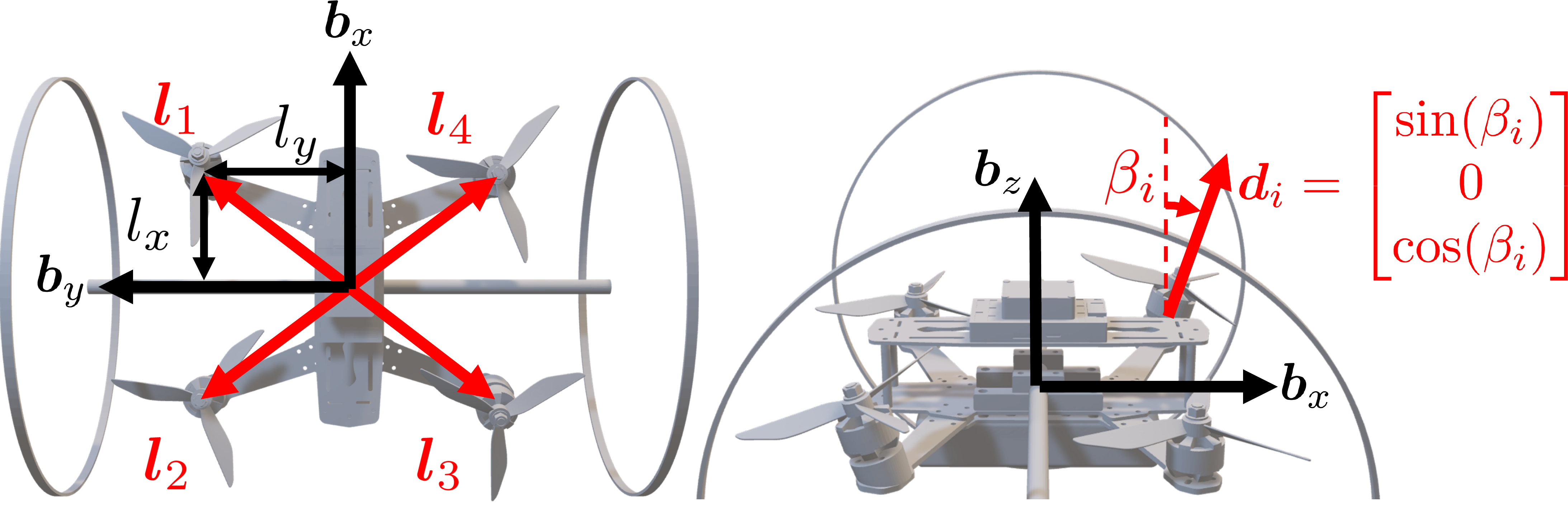}
    \caption{Tilted-rotor configuration of the quadrotor airframe.}
    \label{fig:rotor_tilt}
\end{figure}

We henceforth specialize to the quadrotor configuration shown in
Fig.~\ref{fig:rotor_tilt}: $n_r=4$,
$\bm\ell_1=[\ell_x~\ell_y~0]^\top,\,
\bm\ell_2=[-\ell_x~\ell_y~0]^\top,\,
\bm\ell_3=[-\ell_x~-\ell_y~0]^\top,\,
\bm\ell_4=[\ell_x~-\ell_y~0]^\top,\,
\bm d_i=[\sin\beta_i~0~\cos\beta_i]^\top,\,
\beta_i\in(-\pi/2,\pi/2)$,
i.e., each rotor is tilted about the axle direction by a fixed angle
$\beta_i$; the tilt axes are chosen parallel to $\bm b_y$ so that the rotors
provide force along the rolling direction $\bm b_x$.
With the shorthand $S_i=\sin\beta_i$, $C_i=\cos\beta_i$, direct substitution
into \eqref{eq:wrench_map} yields
\begin{align}
&\bm M=\notag\\
&\begin{bmatrix}
S_1  & S_2  & S_3  & S_4 \\
0 & 0 & 0 & 0\\
C_1  & C_2  & C_3  & C_4 \\
S_{\mk 1\mk}\kappa_{\mk 1\mk}\!+\!C_{\mk 1\mk}\ell_y \!\!&\! S_{\mk 2\mk}\kappa_{\mk 2\mk}\!+\!C_{\mk 2\mk}\ell_y \!\!&\!
S_{\mk 3\mk}\kappa_{\mk 3\mk}\!-\!C_{\mk 3\mk}\ell_y \!\!&\! S_{\mk 4\mk}\kappa_{\mk 4\mk}\!-\!C_{\mk 4\mk}\ell_y\! \\
-C_1\ell_x & C_2\ell_x & C_3\ell_x & -C_4\ell_x \\
C_{\mk 1\mk}\kappa_{\mk 1\mk}\!-\!S_{\mk 1\mk}\ell_y \!\!&\! C_{\mk 2\mk}\kappa_{\mk 2\mk}\!-\!S_{\mk 2\mk}\ell_y \!\!&\!
C_{\mk 3\mk}\kappa_{\mk 3\mk}\!+\!S_{\mk 3\mk}\ell_y \!\!&\! C_{\mk 4\mk}\kappa_{\mk 4\mk}\!+\!S_{\mk 4\mk}\ell_y\!
\end{bmatrix}.
\label{eq:control_effectiveness}
\end{align}
Note that $\bm M_{f,y}=\bm 0$ for this configuration: no lateral thrust is produced. %, which will simplify the friction analysis.

\subsection{Constrained Dynamics}
\label{subsec:constrained_dynamics}

Let the contact surface be the plane
\begin{align}
    \Pi=\{\bm q\in\mbr^3\mid\bm n^\top\bm q=h_f\},
\end{align}
with unit normal $\bm n\in\mathbb{S}^2$ pointing from the surface toward
the vehicle.
Sticking two-wheel contact is characterized by the three conditions
\begin{subequations}
\label{eq:constraints}
\begin{align}
    h_1(\bm p)
    &:=\bm n^\top\bm p-(h_f+r_w)=0,
    \label{eq:contact_constraint}\\
    h_2(\bm p,\dot{\bm p},\bm R)
    &:=\bm b_y^\top\dot{\bm p}=0,
    \label{eq:no_slip_constraint}\\
    h_3(\bm R)
    &:=\bm n^\top\bm b_y=0.
    \label{eq:alignment_constraint}
\end{align}
\end{subequations}
Constraint \eqref{eq:contact_constraint} fixes the distance of the body center from the surface; \eqref{eq:no_slip_constraint} is the nonholonomic no-lateral-slip condition of the wheels; and
\eqref{eq:alignment_constraint} keeps the axle parallel to the plane.
The wheel spin about the axle is a passive internal degree of freedom; rolling resistance and axle friction are assumed negligible.

The admissible velocities under \eqref{eq:constraints} are described as follows.
Define the (world-frame) rolling direction and the body-frame surface normal by $\bm t:=\bm b_y\times\bm n,~\bm n_b:=\bm R^\top\bm n $.
Under \eqref{eq:alignment_constraint}, $\{\bm t,\bm b_y,\bm n\}$ form an orthonormal basis in the world frame, and $\{\bm e_2\times\bm n_b,\bm e_2,\bm n_b\}$ form the corresponding basis in the body frame.
Differentiating \eqref{eq:contact_constraint} gives
$\bm n^\top\dot{\bm p}=0$, which together with
\eqref{eq:no_slip_constraint} confines $\dot{\bm p}$ to the orthogonal
complement of $\mathrm{span}\{\bm n,\bm b_y\}$, namely along $\bm t$. 
Likewise, differentiating \eqref{eq:alignment_constraint} yields
\begin{align*}
   \bm n^\top\dot{\bm b}_y =\bm n^\top\dot{\bm R}\bm e_2=\bm n^\top\bm R(\bm\Omega\times\bm e_2)&=\bm n_b^\top(\bm\Omega\times\bm e_2)\\
    &=\bm\Omega^\top(\bm e_2\times\bm n_b)=0, 
\end{align*}
which implies that $\bm\Omega$ belongs to
$\mathrm{span}\{\bm e_2,\bm n_b\}$.
Therefore, the admissible velocities can be parameterized as
\begin{align}
    \dot{\bm p}=v_t\,\bm t,
    \qquad
    \bm\Omega=\Omega_y\bm e_2+\Omega_n\bm n_b,
    \label{eq:admissible_velocities}
\end{align}
for some $v_t,\Omega_y,\Omega_n\in\mathbb{R}$.
Hence, the non-constrained velocity space is three-dimensional.

Introducing Lagrange multipliers $\lambda_1,\lambda_2,\lambda_3$ for
$h_1,h_2,h_3$ in \eqref{eq:constraints}, the constrained rigid-body dynamics read
\begin{subequations}
\label{eq:constrained_eom}
\begin{align}
    m\ddot{\bm p}
    &=\bm R\bm f-mg\bm e_z
    \!+\!\lambda_1\bm n\!+\!\lambda_2\bm b_y,
    \label{eq:trans_constrained}\\
    \bm J\dot{\bm\Omega}\!+\!\bm\Omega\times\bm J\bm\Omega
    &=\bm\tau
    \!-\!\lambda_3(\bm e_2\times\bm n_b)
    \!-\!r_w\lambda_2(\bm n_b\times\bm e_2).
    \label{eq:rot_constrained}
\end{align}
\end{subequations}
Here $\lambda_1$ is the total normal reaction, $\lambda_2$ the lateral friction force acting along the axle at the contact line (offset $-r_w\bm n$ from the axle center, whence the moment term in \eqref{eq:rot_constrained}), and $\lambda_3$ the alignment-constraint moment transmitted through the differential loading of the two wheels.

%%============================================================================%%
\section{Contact-Mode Hoverability Analysis}
\label{sec:hoverability}

\subsection{Controllability of the Unconstrained State Subspace}

Along any motion satisfying \eqref{eq:constraints}, the projections of
\eqref{eq:constrained_eom} onto $\bm t$, $\bm e_2$, and $\bm n_b$ contain no
Lagrange multiplier. By multiplying \eqref{eq:trans_constrained} by $\bm t^\top$ and \eqref{eq:rot_constrained} by $\bm e_2^\top$ and $\bm n_b^\top$ from the left, we get
\begin{subequations}
\label{eq:projected_eom}
\begin{align}
    m\dot v_t
    &=\bm t^\top\bm R\,\bm M_f\bm u-mg\,\bm t^\top\bm e_z,
    \label{eq:projected_translation}\\
    \bm e_2^\top\!\left(\bm J\dot{\bm\Omega}
        +\bm\Omega\times\bm J\bm\Omega\right)
    &=\bm e_2^\top\bm M_\tau\bm u,
    \label{eq:projected_pitch}\\
    \bm n_b^\top\!\left(\bm J\dot{\bm\Omega}
        +\bm\Omega\times\bm J\bm\Omega\right)
    &=\bm n_b^\top\bm M_\tau\bm u,
    \label{eq:projected_yaw}
\end{align}
\end{subequations}
where $v_t=\bm t^\top\dot{\bm p}$.
Here, we use $\bm t\perp\bm n$, $\bm t\perp\bm b_y$, and $m\,\bm t^\top\ddot{\bm p} =m\frac{d}{dt}(\bm t^\top\dot{\bm p})-m\dot{\bm t}^\top\dot{\bm p} =m\dot v_t$, since $\dot{\bm p}=v_t\bm t$ and $\dot{\bm t}^\top\bm t=0$ ($\|\bm t\|\equiv1$).

Let a nominal contact attitude be $\bm R_\ast\in\SO$ which satisfies $\bm n^\top\bm R_\ast\bm e_2=0$, and define $\bm b_y^\ast=\bm R_\ast\bm e_2$, $\bm t_\ast=\bm b_y^\ast\times\bm n$, $\bm n_b^\ast=\bm R_\ast^\top\bm n$.
The reduced wrench map at $\bm R_\ast$ is
\begin{align}
    \tilde{\bm M}(\bm R_\ast)
    :=
    \begin{bmatrix}
        \bm t_\ast^\top\bm R_\ast\bm M_f\\[2pt]
        \bm e_2^\top\bm M_\tau\\[2pt]
        \bm n_b^\ast{}^\top\bm M_\tau
    \end{bmatrix}
    \in\mbr^{3\times n_r}.
    \label{eq:reduced_wrench_map}
\end{align}
Then, mapping rotor thrusts to the reduced wrench, corresponding to the admissible velocities \eqref{eq:admissible_velocities}, is $\bm w=\tilde{\bm M}(\bm R_\ast)\bm u$.
Let $\bm u_\ast$ be a contact input which keeps the attitude $\bm R_\ast$.
Evaluating \eqref{eq:projected_eom} at a static state
($v_t=0$, $\bm\Omega=\bm 0$) yields the \emph{reduced equilibrium condition}
\begin{align}
    \bm w_{\rm eq}(\bm R_\ast)
    :=\tilde{\bm M}(\bm R_\ast)\,\bm u_\ast
    = \begin{bmatrix}
        mg\,\bm t_\ast^\top\bm e_z\\ 0\\ 0
    \end{bmatrix}.
    \label{eq:reduced_equilibrium}
\end{align}

Let attitude around $\bm R_\ast$ be
$\bm R=\bm R_\ast\exp(\hat{\bm\xi})$ with
$\bm\xi=\eta_y\bm e_2+\eta_n\bm n_b^\ast$, let $e_t$ denote the tangential
displacement, and collect them as
\begin{align}
    \bm q=[e_t~\eta_y~\eta_n]^\top,
    \qquad
    \bm x_T=[\bm q^\top~\dot{\bm q}^\top]^\top\in\mbr^6 .
\end{align}
Then, the linearization of the reduced dynamics \eqref{eq:projected_eom} becomes
\begin{align}
    \dot{\bm x}_T
    =\bm A_T\bm x_T+\bm B_T\delta\bm u,
    \;
    \bm A_T\!=\!
    \begin{bmatrix}
        \bm 0&\bm I_3\\
        \bm G&\bm 0
    \end{bmatrix},
    \;
    \bm B_T\!=\!
    \begin{bmatrix}
        \bm 0\\ \bm H
    \end{bmatrix},
    \label{eq:transverse_linearized}
\end{align}
with
\begin{align}
    \bm H=\bm D\tilde{\bm M}(\bm R_\ast),
    \qquad
    \bm D=\diag\!\left(\tfrac1m,\tfrac1{J_y},\tfrac1{J_n}\right),
\end{align}
where $J_n=(\bm n_b^\ast)^\top\bm J\bm n_b^\ast$, and $\bm G\in\mbr^{3\times3}$ collects the configuration-dependent (thrust--attitude and gravity--attitude) couplings evaluated at
$(\bm R_\ast,\bm u_\ast)$.
For instance at the horizontal contact posture ($\bm n=\bm e_z$, level attitude, $\bm n_b^\ast=\bm e_3$, $J_n=J_z$), a direct computation gives the explicit form
\begin{align}
    \bm G=
    \begin{bmatrix}
        0&f_z^\ast/m&0\\
        0&0&0\\
        0&-\tau_x^\ast/J_z&0
    \end{bmatrix},
    \quad
    \begin{aligned}
        f_z^\ast&=\bm M_{f,z}^\top\bm u_\ast,\\
        \tau_x^\ast&=\bm M_{\tau,x}^\top\bm u_\ast,
    \end{aligned}
    \label{eq:coupling_matrix}
\end{align}
i.e., pitching by $\eta_y$ rotates the equilibrium thrust $f_z^\ast$ into the tangential direction and roll torque $\tau_x^\ast$ into the yaw direction.
The following lemma characterizes controllability.

\begin{lemma}
\label{lm:rank_controllability}
The pair $(\bm A_T,\bm B_T)$ in \eqref{eq:transverse_linearized} is
controllable if and only if
\begin{align}
    \rank\begin{bmatrix}
        \bm H&\bm G\bm H&\bm G^2\bm H
    \end{bmatrix}=3,
    \label{eq:GH_rank}
\end{align}
% i.e., if and only if the pair $(\bm G,\bm H)$ is controllable.
In particular:
\begin{enumerate}
    \item[(i)] if $\rank\tilde{\bm M}(\bm R_\ast)=3$, then
    $(\bm A_T,\bm B_T)$ is controllable for every $\bm G$;
    \item[(ii)] if $\bm G=\bm 0$, then $(\bm A_T,\bm B_T)$ is controllable
    if and only if $\rank\tilde{\bm M}(\bm R_\ast)=3$.
\end{enumerate}
\end{lemma}

\begin{IEEEproof}
From the block structure of $\bm A_T$,
\begin{align*}
    \bm A_T^{2k}\bm B_T&=
    \begin{bmatrix}
        \bm 0\\ \bm G^k\bm H
    \end{bmatrix},\quad
    \bm A_T^{2k+1}\bm B_T=
    \begin{bmatrix}
        \bm G^k\bm H\\ \bm 0
    \end{bmatrix},
    \quad k\ge0.
\end{align*}
Hence the controllability matrix $[\bm B_T~\dots~\bm A_T^5\bm B_T]$ has column space
$\mathcal R\times\mathcal R$, where
$
\mathcal R=\operatorname{Im}
[\bm H~\bm G\bm H~\bm G^2\bm H].
$
Therefore, $(\bm A_T,\bm B_T)$ is controllable if and only if
\eqref{eq:GH_rank} holds. 
Since $\bm D$ is nonsingular, $\rank\bm H=\rank\tilde{\bm M}=3$ gives $\mathcal R=\mbr^3$, then statement (i) holds. Statement (ii) is immediate.
\end{IEEEproof}

A standard non-tilted configuration has a rank-deficient reduced map (see Remark~\ref{rem:tangential_force_tilting}).
When $f_z^\ast\neq0$, the coupling through $\bm G$ can still satisfy \eqref{eq:GH_rank}, yielding the pitch-to-translate motion. By contrast, $\rank\tilde{\bm M}=3$ provides direct actuation of all reduced coordinates as following proposition, guaranteeing controllability without relying on this coupling.
\begin{proposition}[Full reduced actuation]
\label{prop:full_actuation}
The following are equivalent:
\begin{enumerate}
    \item[(i)] $\rank\tilde{\bm M}(\bm R_\ast)=3$;
    \item[(ii)] For any state $\bm q\in\mbr^3$, for any desired $\ddot{\bm q}\in\mbr^3$, there exists
    $\bm u\in\mbr^{n_r}$ such that $\ddot{\bm q}=\bm G\bm q+\bm H\bm u$.
\end{enumerate}
\end{proposition}

\begin{IEEEproof}
Since $\bm D$ is nonsingular,
$\rank\bm H=\rank\tilde{\bm M}$.
Hence $\rank\tilde{\bm M}=3$ is equivalent to the surjectivity of
$\bm H$. Therefore, for any $\bm q,~\ddot{\bm q}\in\mbr^3$, there exists
$\bm u\in\mbr^{n_r}$ satisfying
$\ddot{\bm q}-\bm G\bm q=\bm H\bm u$, which is statement~(ii).
\end{IEEEproof}
The condition in Proposition~\ref{prop:full_actuation} is sufficient for controllability, since controllability may still be achieved through the coupling matrix $\bm G$ even when $\tilde{\bm M}(\bm R_\ast)$ is rank deficient.
Note that Proposition~\ref{prop:full_actuation} only guarantees the existence of a real-valued input $\bm u \in\mbr^{n_r}$. In practice, however, each rotor generates only nonnegative thrust $\bm u \in \mathcal{U}$. This input constraint is taken into account in Section~\ref{ssec:feasibleforceset}.

\subsection{Constraint Satisfaction in the Constrained State Subspace}
\label{subsec:contact_margin}

% Rank conditions concern directions; physical validity of the contact mode concerns magnitudes.
At the nominal attitude, define the quasi-static reactions obtained from the constrained rows of \eqref{eq:trans_constrained}:
\begin{subequations}
\label{eq:static_reactions}
\begin{align}
    \lambda_1^{\rm qs}(\bm u)
    &:=mg\,\bm n^\top\bm e_z
      -\bm n^\top\bm R_\ast\bm M_f\bm u,\\
    \lambda_2^{\rm qs}(\bm u)
    &:=-(\bm b_y^\ast)^\top\!\left(\bm R_\ast\bm M_f\bm u-mg\bm e_z\right).
\end{align}
\end{subequations}
Since $\lambda_1^{\rm qs}$ and $\lambda_2^{\rm qs}$ are the normal reaction and the lateral friction, respectively, they are physically valid only if
\begin{align}
    \lambda_1^{\rm qs}\ge0,
    \qquad
    |\lambda_2^{\rm qs}|\le\mu\lambda_1^{\rm qs},
    \label{eq:friction_condition}
\end{align}
with friction coefficient $\mu>0$.
The moment $\lambda_3$ is transmitted through differential wheel loading and is bounded by the requirement that both wheels stay in contact. Let $n_1,n_2\ge0$ denote the normal loads of the two wheels, located at $\mp\ell_w\bm e_2$ from the body center. Force and moment balance about the rolling axis give
\begin{align}
    n_1+n_2=\lambda_1^{\rm qs},
    \quad
    n_{1,2}
    =\frac12\left(\lambda_1^{\rm qs}\pm\frac{\tau_{t_b}(\bm u)}{\ell_w}\right),
    \label{eq:wheel_loads}
\end{align}
where $\tau_{t_b}(\bm u)=(\bm t\bm R^\top)^\top\bm M_\tau\bm u$ is the roll torque about
the rolling axis in the body frame.
Nonnegativity of both loads is therefore equivalent to
$\lambda_1^{\rm qs}\ge|\tau_{t_b}/\ell_w|$.
Aggregating unilaterality, sticking, and two-wheel loading, define the
\emph{quasi-static contact margin}
\begin{align}
    m_c^{\rm qs}(\bm u)
    :=\min
    \left\{
        \lambda_1^{\rm qs},\,
        \mu\lambda_1^{\rm qs}-|\lambda_2^{\rm qs}|,\,
        \lambda_1^{\rm qs}-\left|\frac{\tau_{t_b}}{\ell_w}\right|
    \right\}.
    \label{eq:contact_margin}
\end{align}
% where the last entry is normalized by $\ell_w$ so that all entries carry the unit of force.

\begin{lemma}
\label{lem:contact_condition}
If $m_c^{\rm qs}(\bm u)\geq0$, then the unilateral contact condition, the lateral sticking condition, and positive loading of both wheels are all satisfied at $\bm R_\ast$.
\end{lemma}

\begin{IEEEproof}
    This follows immediately from \eqref{eq:friction_condition} and \eqref{eq:wheel_loads}.
\end{IEEEproof}
The contact constraint allows a nonzero torque $\tau_{t_b}$ as long as it can be supported by the differential wheel loading. This enables the vehicle to generate tangential force while using the wheel--surface contact.

\subsection{Contact-Mode Hoverability}
\label{subsec:hoverability_lp}

In free flight, \emph{hoverability} was defined in \cite{Ito19052023} as the existence of a nonnegative equilibrium input for which the linearization is controllable. We adapt this concept to the contact-mode.

\begin{definition}[Contact-mode hoverability]
\label{def:hoverability}
A two-wheeled drone is \emph{contact-mode hoverable} at $\bm R_\ast$ if
there exists $\bm u_\ast\in\mathrm{int}\,\mathcal U$ such that
\begin{enumerate}
    \item[(i)] the reduced equilibrium \eqref{eq:reduced_equilibrium} holds;
    \item[(ii)] the contact condition $m_c^{\rm qs}(\bm u_\ast)>0$ holds;
    \item[(iii)] the linearized reduced dynamics \eqref{eq:transverse_linearized} are controllable.
\end{enumerate}
\end{definition}

\begin{definition}[Hoverability margin]
    \emph{Hoverability margin} $\gamma^\star$ is defined as the optimal value of the linear
    program
    \begin{subequations}
    \begin{align}
        \gamma^\star =&~\max_{\bm u,\,\gamma}~\gamma \label{eq:hoverability_lp} \\
        \text{subject to}~&\tilde{\bm M}(\bm R_\ast)\bm u =\bm w_{\rm eq}(\bm R_\ast), \label{eq:hover_lp_eq}\\
        &\gamma\le u_i,~\gamma\le f_{\max}\!-\!u_i,~i=1,\ldots,n_r, \label{eq:hover_lp_input}\\
        &\gamma\le w_{\rm qs} m_c^{\rm qs}(\bm u). \label{eq:hover_lp_contact}
    \end{align}
    \end{subequations}
    where $w_{\rm qs}$ is a weighting parameter.
\end{definition}

\begin{theorem}[Sufficient condition for contact-mode hoverability]
\label{thm:hoverability_lp}
If hoverability margin $\gamma^\star>0$ and $\rank\tilde{\bm M}(\bm R_\ast)=3$, then the vehicle
is contact-mode hoverable at $\bm R_\ast$, with actuator input margins
at least $\gamma^\star$.
\end{theorem}

\begin{IEEEproof}
Constraint \eqref{eq:hover_lp_eq} enforces (i);
\eqref{eq:hover_lp_input} with $\gamma^\star>0$ gives $\bm u\in\mathrm{int}\,\mathcal U$; \eqref{eq:hover_lp_contact} with $\gamma^\star>0$ gives (ii).
The rank condition yields (iii) by Lemma~\ref{lm:rank_controllability}(i), for every coupling $\bm G$.
\end{IEEEproof}

%%============================================================================%%
\section{Structure Optimization for Horizontal Locomotion}
\label{sec:tilt_optimization}

Although the preceding analysis holds for arbitrary structures and contact planes, we focus here on optimizing the vehicle configuration for horizontal floor locomotion. 
In this case, $\bm t_\ast^\top\bm R_\ast$ selects $\bm b_x$ and $\bm n_b^\ast=\bm e_3$.

%%============================================================================%%
\subsection{Symmetric Rotor-Tilt Designs}
\label{sec:tilt_optimization}

The following observation reveals a fundamental limitation of the standard configuration.
\begin{remark}
\label{rem:tangential_force_tilting}
For a standard non-tilted configuration ($\beta_i = 0,~i=1,\dots,4$), the first row of \eqref{eq:control_effectiveness} is zero, yielding $\rank \tilde{\bm M}(\bm R_\ast) \le 2$ and $\bar f_t = 0$.
Hence, tangential acceleration under level contact relies entirely on the attitude coupling $\bm G$ (i.e., body pitching). In contrast, a tilted-rotor configuration ($\bm M_{f,x} \neq \bm 0$) generates tangential force directly without body pitch.
\end{remark}

Hereafter, we consider two symmetric rotor-tilt layouts, one of which will be shown to be rank-deficient.

\begin{definition}[Mirror symmetry structure]
Consider the layout symmetric with respect to the $b_y$--$b_z$ plane,
\begin{align}
    \beta_2=-\beta_1,~\kappa_2=-\kappa_1,
    \qquad
    \beta_4=-\beta_3,~\kappa_4=-\kappa_3.
    \label{eq:mirror_symmetry}
\end{align}
\end{definition}

\begin{proposition}
\label{prop:mirror}
Under \eqref{eq:mirror_symmetry}, $\rank\tilde{\bm M}(\bm R_\ast)\le2$ for all tilt angles.
\end{proposition}

\begin{IEEEproof}
Substituting $\bm t_\ast^\top\bm R_\ast=\bm b_x$, $\bm n_b^\ast=\bm e_3$, and \eqref{eq:mirror_symmetry} into
\eqref{eq:control_effectiveness} and \eqref{eq:reduced_wrench_map} yields
\begin{align*}
    &\tilde{\bm M}(\bm R_\ast)=\\
    &\begin{bmatrix}
        S_1&-S_1&S_3&-S_3\\
        -\ell_xC_1&\ell_xC_1&\ell_xC_3&-\ell_xC_3\\
        \kappa_1C_1\!-\!\ell_yS_1 \!\!&\! -\kappa_1C_1\!+\!\ell_yS_1 \!\!&\!
        \kappa_3C_3\!+\!\ell_yS_3 \!\!&\! -\kappa_3C_3\!-\!\ell_yS_3
    \end{bmatrix},
\end{align*}
whose second and fourth columns are the negatives of the first and third, respectively. The column space is spanned by two vectors.
\end{IEEEproof}

\begin{definition}[Half-turn symmetry structure]
Consider the layout invariant under a half-turn about $\bm b_z$,
\begin{align}
    \beta_3=-\beta_1,~\kappa_3=\kappa_1,
    \qquad
    \beta_4=-\beta_2,~\kappa_4=\kappa_2.
    \label{eq:halfturn_symmetry}
\end{align}
\end{definition}

\begin{proposition}
\label{prop:halfturn}
Under \eqref{eq:halfturn_symmetry},
\begin{align}
    \rank\tilde{\bm M}(\bm R_\ast)=3
    \iff
    \begin{cases}
        \sin(\beta_1+\beta_2)\neq0,\\[2pt]
        (a_1,a_2)\neq(0,0),
    \end{cases}
    \label{eq:halfturn_condition}
\end{align}
where $a_i:=\kappa_iC_i-\ell_yS_i$, $i=1,2$, is the reduced yaw authority of
rotor $i$.
\end{proposition}

\begin{IEEEproof}
Substitution of \eqref{eq:halfturn_symmetry} gives
\begin{align}
    \tilde{\bm M}(\bm R_\ast)=
    \begin{bmatrix}
        S_1&S_2&-S_1&-S_2\\
        -\ell_xC_1&\ell_xC_2&\ell_xC_1&-\ell_xC_2\\
        a_1&a_2&a_1&a_2
    \end{bmatrix}.
    \label{eq:E_halfturn}
\end{align}
The column space is unchanged by replacing the columns
$(\bm c_1,\bm c_2,\bm c_3,\bm c_4)$ with
$(\bm c_1\!-\!\bm c_3,\,\bm c_2\!-\!\bm c_4,\,
  \bm c_1\!+\!\bm c_3,\,\bm c_2\!+\!\bm c_4)/2$, i.e., with
\begin{align*}
    \begin{bmatrix}S_1\\-\ell_xC_1\\0\end{bmatrix},~
    \begin{bmatrix}S_2\\\ell_xC_2\\0\end{bmatrix},~
    \begin{bmatrix}0\\0\\a_1\end{bmatrix},~
    \begin{bmatrix}0\\0\\a_2\end{bmatrix}.
\end{align*}
The first two span the $(f_t,\tau_y)$-plane iff their $2\times2$ determinant $\ell_x(S_1C_2+C_1S_2)=\ell_x\sin(\beta_1+\beta_2)$ is nonzero; the last two span the $\tau_n$-axis iff some $a_i\neq0$.
Rank $3$ holds iff both hold; if either fails the column space lies in a plane.
\end{IEEEproof}

Equation~\eqref{eq:halfturn_condition} reveals two structural limitations: the untilted configuration $\beta_1=\beta_2=0$ violates $\sin(\beta_1+\beta_2)\neq0$ (recovering Remark~\ref{rem:tangential_force_tilting}), and the yaw authority $a_i=\kappa_i\cos\beta_i-\ell_y\sin\beta_i$ vanishes at $\tan\beta_i=\kappa_i/\ell_y$, due to the counter-torque and tilted-thrust moments cancellation.

Under \eqref{eq:halfturn_symmetry}, the admissible rotor-spin patterns reduce, up to relabeling, to
\begin{align}
    \text{4P:}~\kappa_1=\kappa_2=\kappa,
    \qquad
    \text{PNPN:}~\kappa_1=-\kappa_2=\kappa,
    \label{eq:spin_patterns}
\end{align}
Thus, the spins are either identical or alternating. The design question is which pattern and tilt angles maximize traction.

\subsection{Feasible Wrench Set} \label{ssec:feasibleforceset}
% Rank conditions certify directions of actuation; design requires quantifying magnitudes.
Under the contact constraints, Lemma \ref{lem:contact_condition}, the \emph{feasible reduced-wrench set} is defined as the convex polytope
\begin{align}
    \mathcal W(\bm R_\ast)
    =\left\{
        \tilde{\bm M}(\bm R_\ast)\bm u~\middle|~
        \bm u\in\mathcal U,~m_c^{\rm qs}(\bm u)\ge0
    \right\},
    \label{eq:wrench_polytope}
\end{align}
the image of a polytope under a linear map.
For locomotion, the most relevant scalar is the tangential force sustainable without disturbing the attitude equilibrium. We therefore define the maximum traction force
\begin{align}
    \bar f_t
    :=\max\big\{\,\bm e_1^\top\bm w~\big|~
        \bm w\in\mathcal W(\bm R_\ast),~
        \bm e_2^\top\bm w\!=\!\bm e_3^\top\bm w\!=\!0
    \,\big\}.
    \label{eq:level_cruise_traction}
\end{align}
% the support of the section $\mathcal W\cap\{\tau_y=\tau_n=0\}$.
% By Proposition~\ref{prop:full_actuation}, $\bar f_t>0$ requires the
% tangential force row to act nontrivially on the null space of the reduced
% torque rows.

\begin{proposition}
    \label{rem:Fset}
    The optimization problem in \eqref{eq:level_cruise_traction} is equivalent to maximizing $f_x$ over the set $\mathcal F=\mathcal F_c\cap\mathcal F_{nc}$ with
    \begin{subequations}
    \begin{align}
        \mathcal F_c&=
        \left\{[\ast~f_z~\tau_x]^\top~\middle|~
            \ell_w(mg-f_z)\ge|\tau_x|\right\},
        \label{eq:feasible_set_constrained}\\
        \mathcal F_{nc}&=
        \left\{
        \begin{bmatrix}
            f_x\\ f_z\\ \tau_x
        \end{bmatrix}
        = \begin{bmatrix}
            \bm M_{f,x}^\top\\ \bm M_{f,z}^\top\\ \bm M_{\tau,x}^\top
        \end{bmatrix}\bm u~\middle|~
        \bm u\in\mathcal U\cap
        \Null\!\begin{bmatrix}
            \bm M_{\tau,y}^\top\\ \bm M_{\tau,z}^\top
        \end{bmatrix}
        \right\},
        \label{eq:feasible_set_unconstrained}
    \end{align}
    \end{subequations}
\end{proposition}
\begin{IEEEproof}
    The conditions in \eqref{eq:level_cruise_traction} are $\bm u\in\mathcal U$, $\bm e_5^{(6)\top}\!\bm w=\bm e_6^{(6)\top}\!\bm w=0$, and $m_c^{\rm qs}(\bm u)\ge0$. The first two conditions are equivalent to $\bm u \in \mathcal{U} \cap \Null [\bm M_{\tau,y}~\bm M_{\tau,z}]^\top$.
    Substituting the horizontal-attitude conditions $\bm t_\ast^\top\bm R_\ast=\bm b_x$ and $\bm n_b^\ast=\bm e_3$ into \eqref{eq:contact_margin}, together with \eqref{eq:static_reactions}, yields
    \begin{align}
        &m_c^{\rm qs}(\bm u)\ge0 \notag\\
        &\Leftrightarrow
        \min \left\{ mg\!-\!f_z,~\mu(mg\!-\!f_z),~ (mg\!-\!f_z) \!-\! \frac{|\tau_x|}{\ell_w} \right\} \ge 0 \notag\\
        &\Leftrightarrow \ell_w(mg-f_z)\ge|\tau_x|.
        \label{eq:conditions}
    \end{align}
    Thus, \eqref{eq:level_cruise_traction} is equivalent to maximizing $f_x$ subject to \eqref{eq:conditions}. Combining these conditions yields $\mathcal F=\mathcal F_c\cap\mathcal F_{nc}$.
\end{IEEEproof}
The set $\mathcal F$ offers a convenient three-dimensional visualization in
the traction--load--roll space
(Fig.~\ref{fig:feasible_wrench}).

\subsection{Configuration and Tilt-Angle Optimization}
\label{subsec:tilt_design}

Let $\beta=(\beta_1,\beta_2)$ be the independent tilt variables under
half-turn symmetry.
We select the tilt angles by solving
\begin{align}
    \label{eq:tilt_design}
    \beta^\ast =\argmin_{\beta}& -\bar f_t(\beta) -w_\gamma \gamma^\star(\beta) +w_\beta\|\beta\|^2 \notag\\
    & +w_c\left\{\cond\!\big(\bm M_\mathrm{hov}(\beta)\big)+\cond\!\big(\bm M_\mathrm{gnd}(\beta)\big)\right\} \\
    \mathrm{subject~to}&~ \gamma^\star(\beta) \geq \varepsilon_\gamma \\
    &~\rank\tilde{\bm M}(\bm R_\ast)(\beta)=3
\end{align}
where $\bar f_t(\beta)$ is the level-cruise traction \eqref{eq:level_cruise_traction}, $\gamma^\star(\beta)$ is the LP hoverability margin \eqref{eq:hoverability_lp}, and $\cond(\cdot)$ denotes the condition number.
The first term maximizes direct traction at a level attitude; the second rewards contact-margin robustness; the third penalizes large tilt angles, which sacrifice vertical thrust efficiency in the flight mode; the fourth keeps the hovering and ground-mode allocation matrices well conditioned. This condition safeguards the flight-mode allocation of Section~\ref{sec:controller}, which inverts following submatrices of $\bm M$
\begin{align}
    M_\mathrm{hov}(\beta)=\begin{bmatrix}
        \bm M_{f,z}^\top\\ \bm M_\tau
    \end{bmatrix},\quad
    M_\mathrm{gnd}(\beta)=\begin{bmatrix}
        \bm M_{f,x}^\top\\ \bm M_{f,z}^\top\\
        \bm M_{\tau,y}^\top\\ \bm M_{\tau,z}^\top
    \end{bmatrix}.
\end{align}
The constraints guarantee the contact-mode hoverability according to Theorem \ref{thm:hoverability_lp}.

For a fair comparison, the same model parameters and optimization settings were used for both the 4P and PNPN configurations, with only the rotor rotation directions being changed. The vehicle mass and gravitational acceleration were set to $m=1.10\,\mathrm{kg}$ and $g=9.81\,\mathrm{m/s^2}$, respectively. The maximum thrust of each rotor was $f_{\max}=4.78\,\mathrm{N}$, and the thrust-to-torque coefficient was $\kappa=0.0153\,\mathrm{m}$. The rotor positions were defined by $\ell_x=0.075\,\mathrm{m}$ and $\ell_y=0.10\,\mathrm{m}$, and the wheel half-length was $\ell_w=0.19\,\mathrm{m}$. The rotor tilt angles were constrained to $-90^\circ \leq \beta_i \leq 90^\circ$. The objective-function weights were set to $w_\beta=100$, $w_{\rm qs}=1$, $w_{\mathrm{cond}}=1$, and $w_\gamma=1$, with $\epsilon_\gamma=0.1$.

\begin{figure}[t]
\centering
\subfloat[]{\includegraphics[trim=0 0 0 0,width=0.5\linewidth,clip]{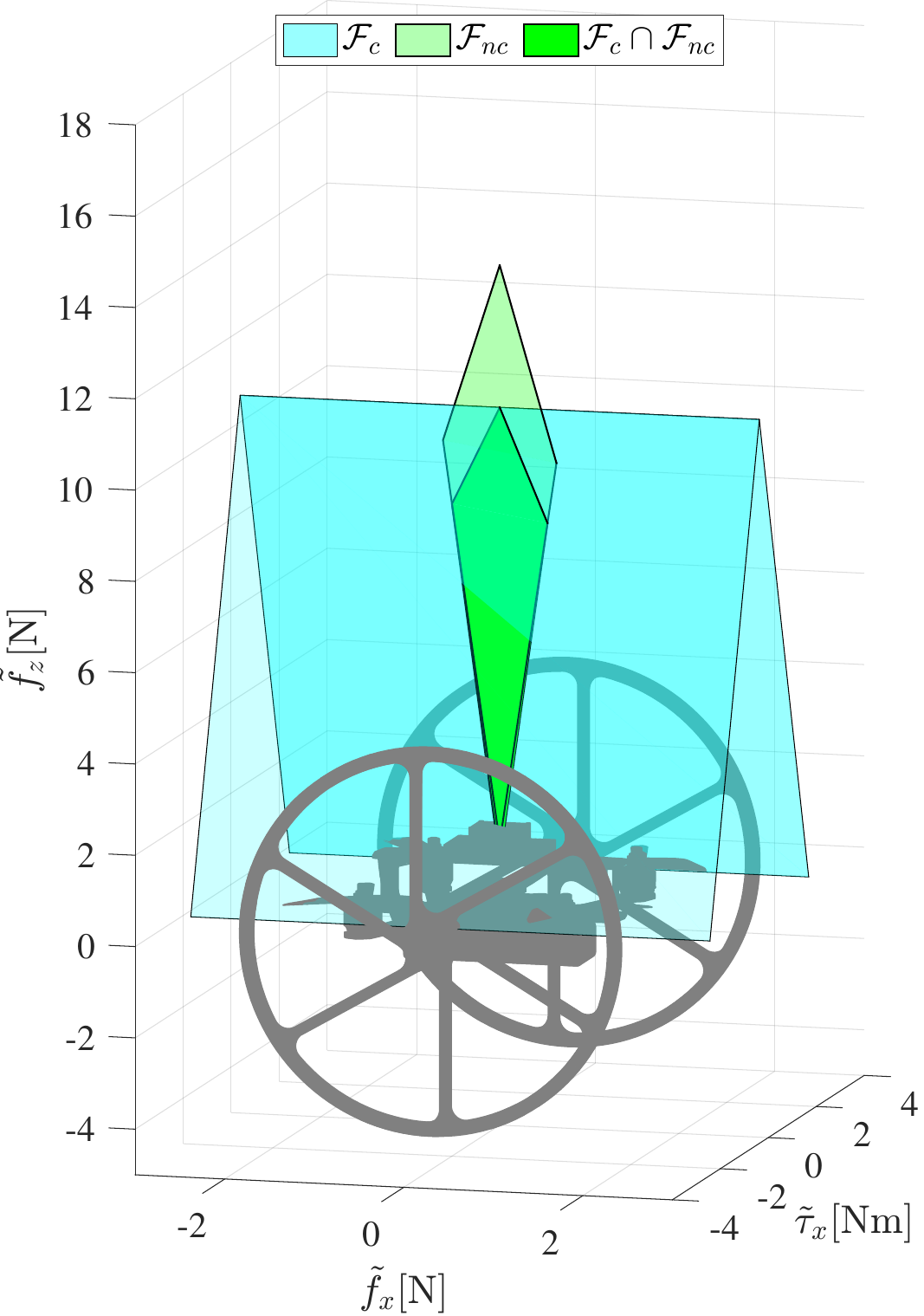}%
\label{fig:feasible_wrench_PNPN}}
\hfil
\subfloat[]{\includegraphics[trim=0 0 0 0,width=0.5\linewidth,clip]{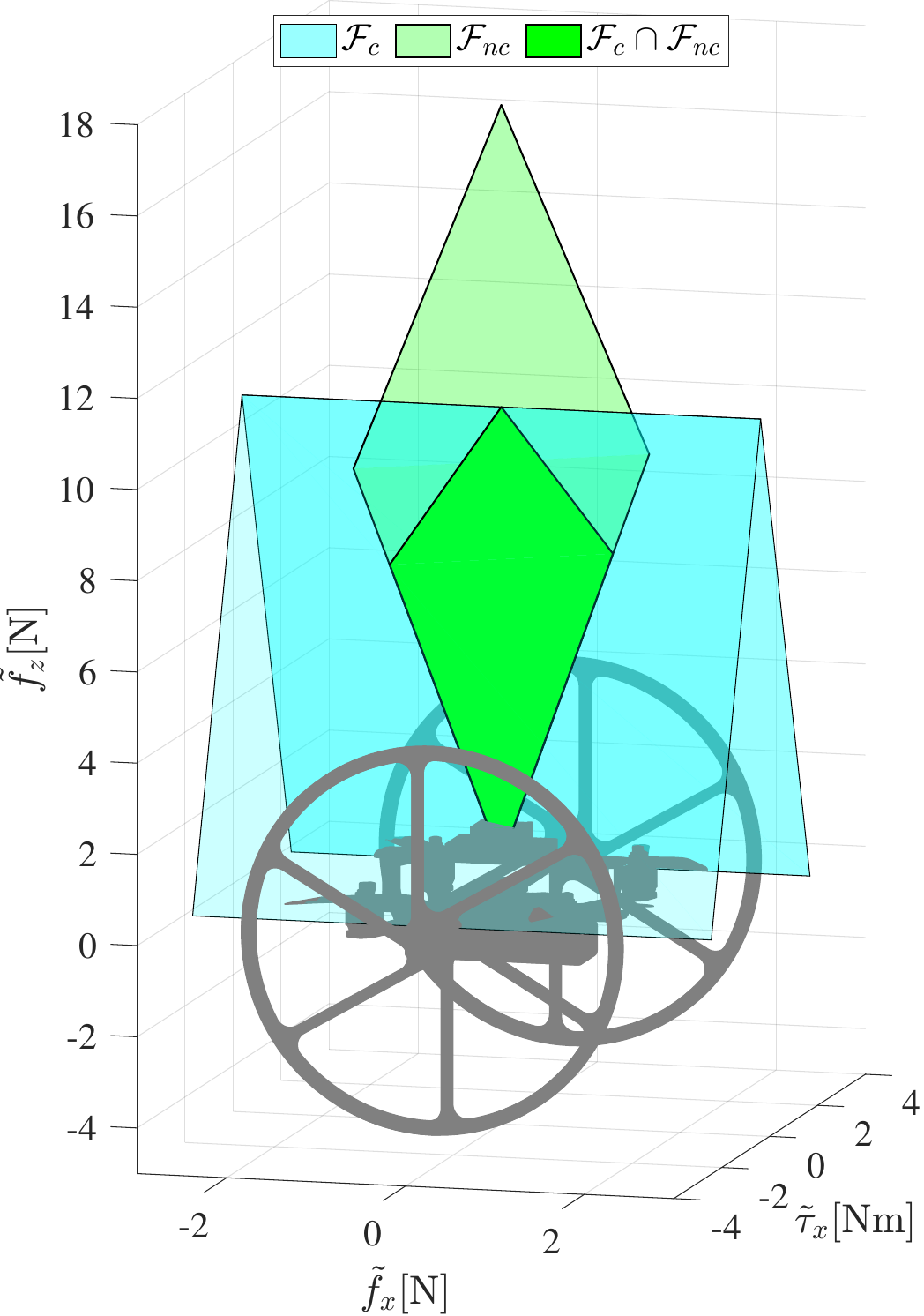}%
\label{fig:feasible_wrench_4P}}
\caption{Feasible wrench sets $\mathcal{F}$ at the optimized rotor tilt angles. (a) PNPN configuration. (b) 4P configuration.}
\label{fig:feasible_wrench}
\end{figure}

The optimized rotor tilt angles were $\boldsymbol{\beta}=[-9.74^\circ,\,28.28^\circ]$ for 4P and $\boldsymbol{\beta}=[-21.16^\circ,\,5.29^\circ]$ for PNPN. Both configurations satisfied the feasibility condition with $\mathrm{rank}(\tilde{\mathbf{M}})=3$. The 4P configuration achieved a larger maximum tangential force, $\bar{f}_t=1.205\,\mathrm{N}$, than PNPN, $\bar{f}_t=0.751\,\mathrm{N}$, as also illustrated by the greater extent of $\mathcal{F}$ in the $f_x$ direction in Fig.~\ref{fig:feasible_wrench}. 
The feasible wrench set $\mathcal F$  represents the wrenches that can be generated by the rotors while satisfying the contact constraints. In particular, tangential force can be generated as long as the accompanying roll torque satisfies $|\tau_x|\leq \ell_w(mg-f_z)$.
The hoverability margin $\gamma^\star$ was $2.210$ for 4P and $1.539$ for PNPN, while the corresponding condition numbers were $\cond(\bm M_\mathrm{hov})=\cond(\bm M_\mathrm{gnd})=28.28$ and $\cond(\bm M_\mathrm{hov})=\cond(\bm M_\mathrm{gnd})=25.45$, respectively. Thus, although PNPN exhibited a slightly lower condition number, 4P achieved larger values of both $\bar{f}_t$ and $\gamma^\star$ while satisfying the required controllability condition.

\begin{figure}[t]
\centering
\subfloat[]{\includegraphics[trim=25 15 50 0,width=0.5\linewidth,clip]{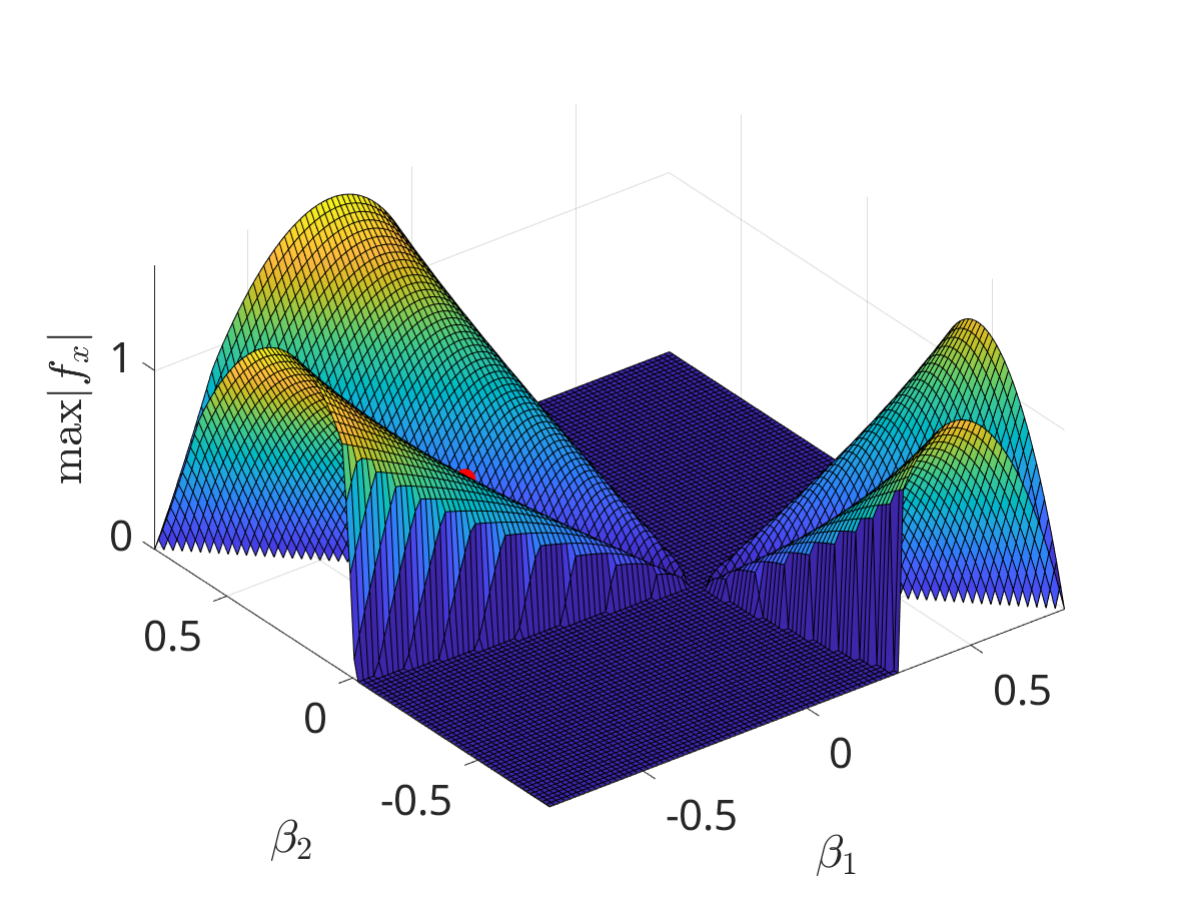}%
\label{fig:fxmax_map_PNPN}}
\hfil
\subfloat[]{\includegraphics[trim=35 5 50 30,width=0.5\linewidth,clip]{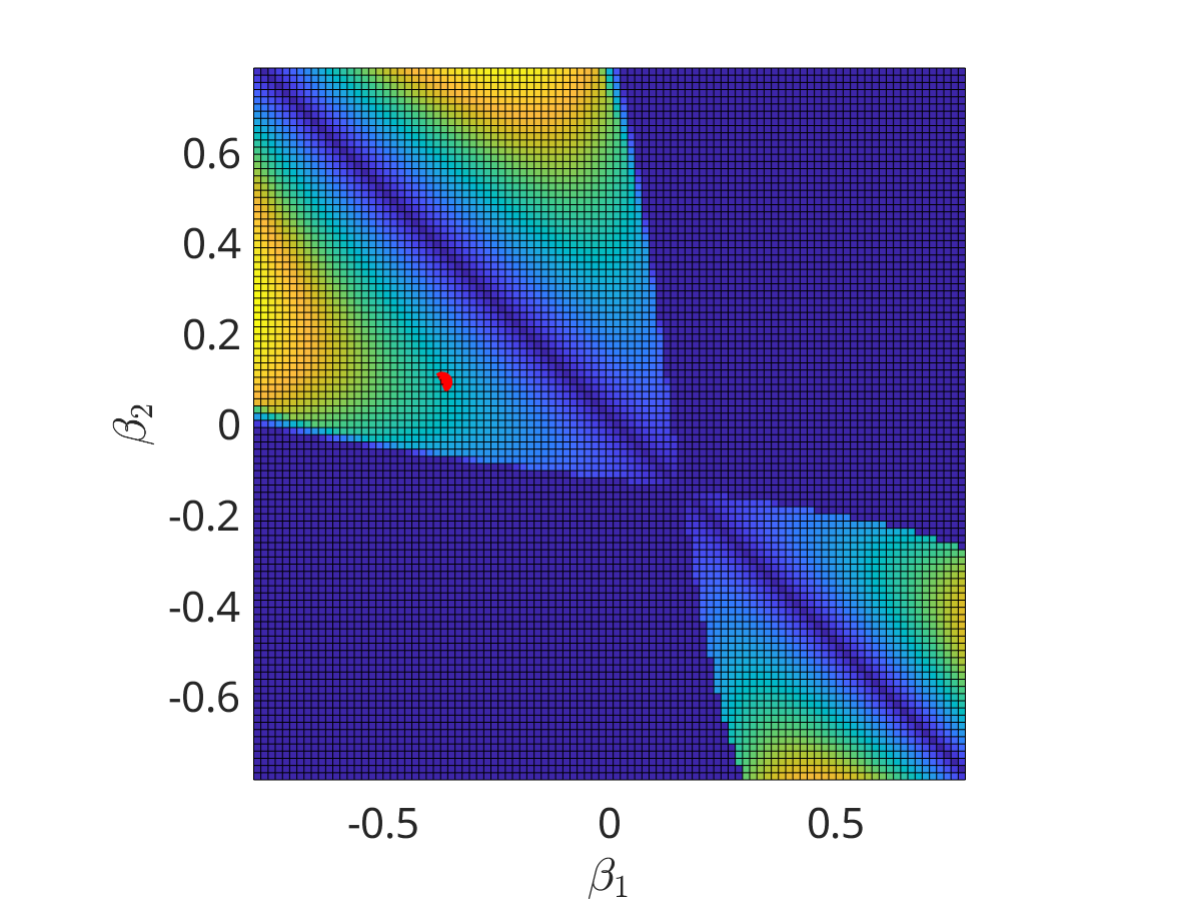}%
\label{fig:fxmax_map_PNPN_top}}
\hfil
\subfloat[]{\includegraphics[trim=25 15 50 0,width=0.5\linewidth,clip]{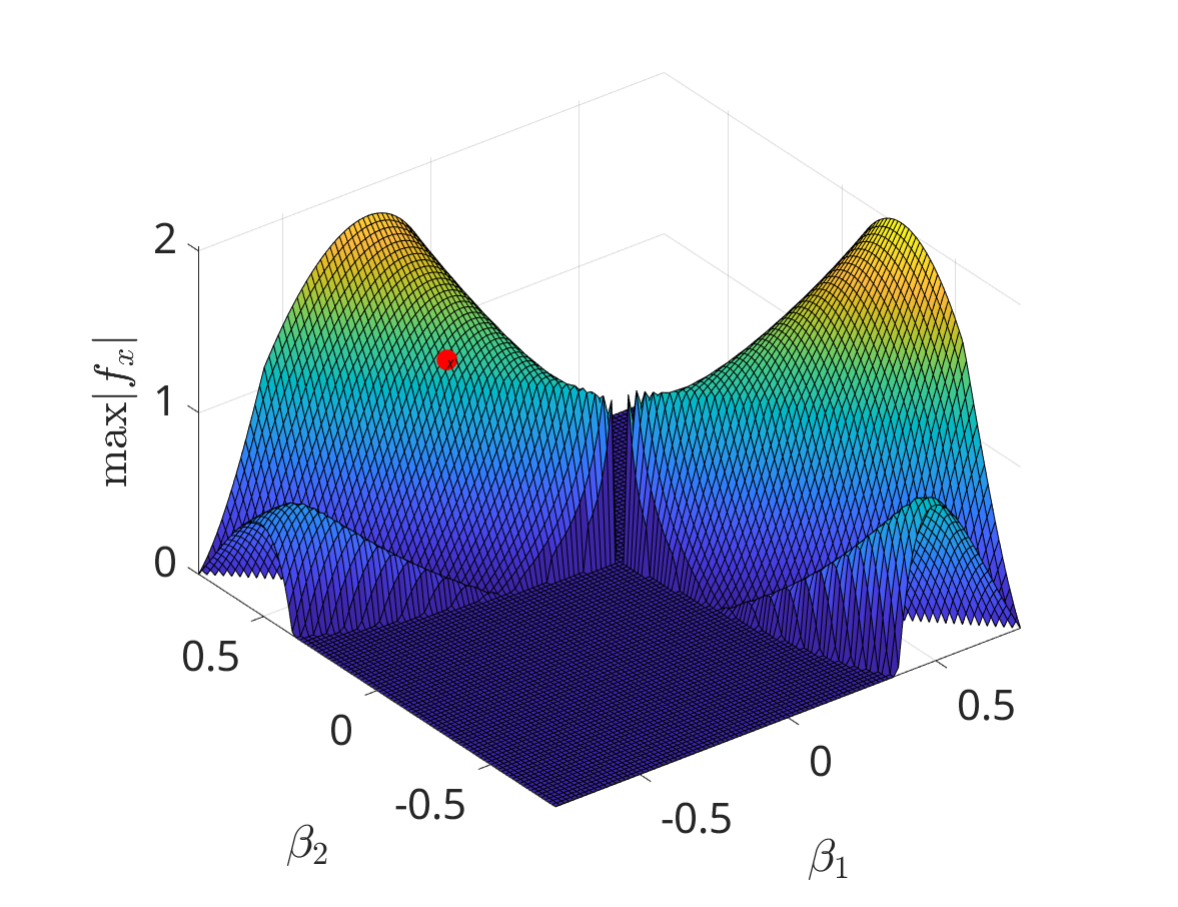}%
\label{fig:fxmax_map_4P}}
\hfil
\subfloat[]{\includegraphics[trim=35 5 50 30,width=0.5\linewidth,clip]{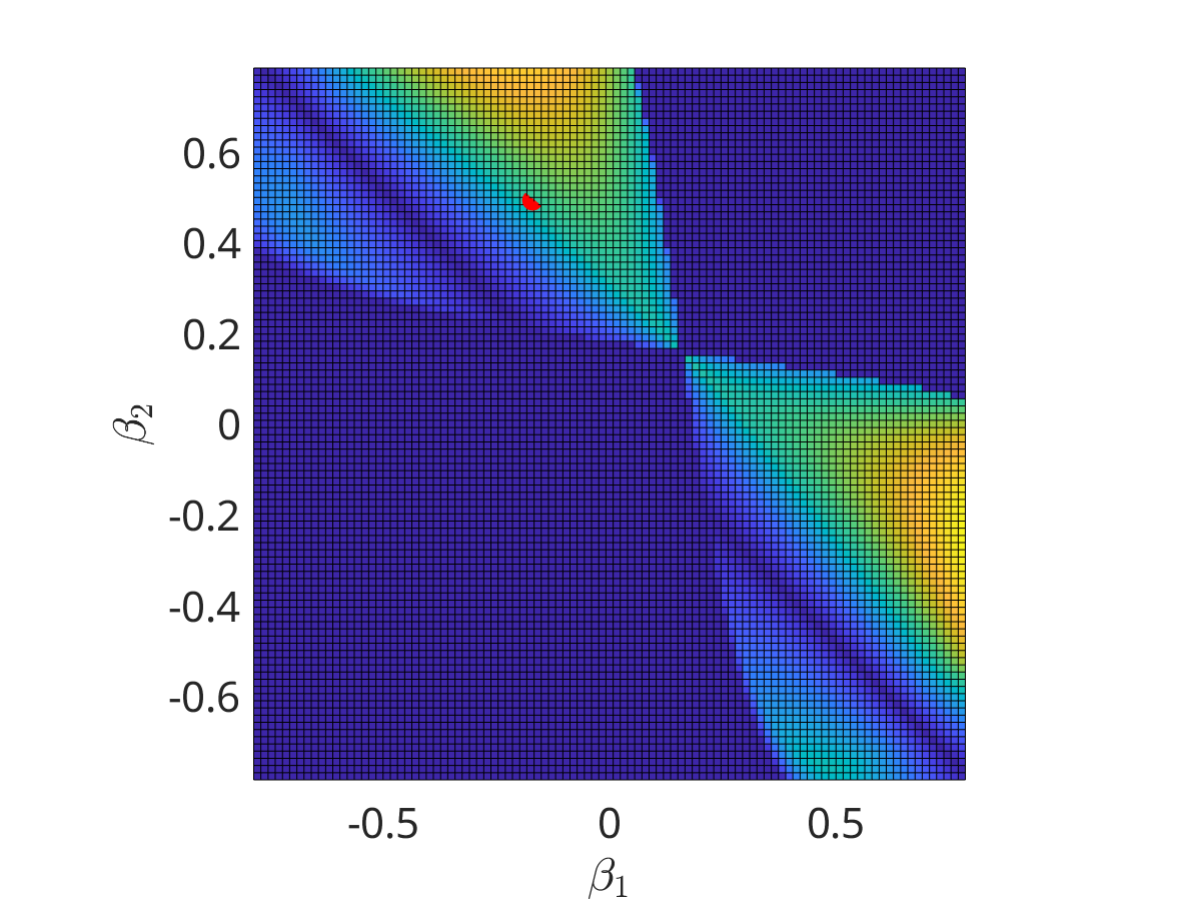}%
\label{fig:fxmax_map_4P_top}}
\caption{Maximum tangential force $\bar{f}_t$ over the rotor tilt-angle plane. The red dots indicate the optimized configurations. (a)-(b) PNPN configuration: side and top view. (c)-(d) 4P configuration: side and top view.}
\label{fig:fxmax_map}
\end{figure}

The dependence of $\bar{f}_t$ on the rotor tilt angles is further shown in Fig.~\ref{fig:fxmax_map}. In addition to achieving a larger $\bar{f}_t$ at the optimized point, 4P maintains relatively large values over a wider range of small tilt angles than PNPN, indicating that its improved tangential force capability is not limited to a specific set of optimized angles. These results show that the rotor rotation configuration significantly affects the feasible wrench capability and that 4P provides greater tangential force capability with relatively small rotor tilt angles. Based on these numerical results, the 4P configuration was selected for the proposed vehicle.

\section{Controller Design}
\label{sec:controller}

This section compares the proposed 4P configuration optimized in Section~\ref{subsec:tilt_design} with an untilted PNPN configuration used as a standard wheeled-drone baseline.
The same controller structure is used for both configurations to evaluate the effect of the rotor configuration.

\subsection{Hovering Controller}

For hovering, we employ a geometric controller on $\SO$ \cite{lee2010}.
Given a reference position $\bm p_d$, the desired force is computed by the
PD law
\begin{align}
    \bm f_d=k_p(\bm p_d-\bm p)-k_v\dot{\bm p}+mg\bm e_z,
    \label{eq:force_input_hov}
\end{align}
and the thrust magnitude by projection onto the current body $z$-axis,
$f_{zd}=\bm f_d^\top\bm b_z$.
With a desired heading direction $\bm b_{xd}$, the desired attitude is
\begin{align}
    \bm R_d=[\,\bm b_{yd}\times\bm b_{zd}~~\bm b_{yd}~~\bm b_{zd}\,],
    \label{eq:desired_att_hov}
\end{align}
where $\bm b_{zd}=\bm f_d/\|\bm f_d\|$ and
$\bm b_{yd}=\bm b_{zd}\times\bm b_{xd}
/\|\bm b_{zd}\times\bm b_{xd}\|$.
The torque input is the PD-type attitude law
\begin{align}
    \bm\tau_d
    =\tfrac12 k_R\big(\bm R^\top\bm R_d-\bm R_d^\top\bm R\big)^\vee
    -k_\Omega\bm\Omega
    +\bm\Omega\times\bm J\bm\Omega,
    \label{eq:torque_input_hov}
\end{align}
and the rotor thrusts are allocated by
\begin{align}
    \bm u=
     M_\mathrm{hov}^{-1}
    \begin{bmatrix}
        f_{zd}\\ \bm\tau_d
    \end{bmatrix}.
    \label{eq:mixer_hov}
\end{align}

\subsection{Ground-Driving Controllers}

For ground driving, gravity compensation is unnecessary.
To maximize the admissible horizontal force, the desired vertical force is
held at the constant $\bar f_z$ at which the tangential section of the
feasible wrench set is widest
(Fig.~\ref{fig:feasible_wrench}\subref{fig:feasible_wrench_4P}), and the
desired force is
\begin{align}
    \bm f_d=k_x(\bm p_d-\bm p)-k_v\dot{\bm p}+\bar f_z\bm e_z,
    \label{eq:force_input_ground}
\end{align}
whose vertical error components vanish on the constraint manifold.
The desired heading points toward the target for forward motion and away
from it for reversing:
\begin{align}
    \bm b_{xd}=
    \begin{cases}
        \dfrac{\bm p_d-\bm p}{\|\bm p_d-\bm p\|},
        &(\bm p_d-\bm p)^\top\bm R_{wa}\bm e_1\ge0,\\[8pt]
        \dfrac{\bm p-\bm p_d}{\|\bm p-\bm p_d\|},
        &\text{otherwise},
    \end{cases}
    \label{eq:desired_x}
\end{align}
where $\bm R_{wa}\in\SO$ is the heading (axle-frame) rotation about
$\bm e_z$; near the target, where the line-of-sight direction is
ill-conditioned, the desired heading is frozen to avoid chattering.

\subsubsection{Standard wheeled drone}
The standard configuration must tilt its body to produce a horizontal force
(Remark~\ref{rem:tangential_force_tilting}).
Its ground controller therefore mirrors the hovering controller:
$f_{zd}$ is computed by projecting \eqref{eq:force_input_ground} onto
$\bm b_z$, the desired attitude follows \eqref{eq:desired_att_hov}, the
torque follows \eqref{eq:torque_input_hov} with $\tau_{dx}=0$ (roll is
carried by the wheels), and the allocation is \eqref{eq:mixer_hov}.

\subsubsection{Proposed 4P wheeled drone}
The proposed structure generates horizontal force directly and thus holds a
level attitude throughout:
\begin{align}
    \bm R_d=[\,\bm b_{xd}~~\bm e_z\times\bm b_{xd}~~\bm e_z\,].
\end{align}
The force input is decoupled into the constant vertical component
$\bar f_z$ and the tangential component
\begin{align}
    f_{xd}=\mathrm{clip}_{[-\eta\bar f_t,\,\eta\bar f_t]}
    \big(\bm f_d^\top\bm R_{wa}\bm e_1\big),
    \label{eq:fxd_clip}
\end{align}
where $\bar f_t$ is the level-cruise traction defined in
\eqref{eq:level_cruise_traction}. Since $\bar f_t$ lies on the boundary
of the wheel-loading constraint, we use $\eta=0.95$ to preserve a
positive contact margin under transient attitude corrections.
With $\bm\tau_d$ from \eqref{eq:torque_input_hov}, the thrusts are allocated by
\begin{align}
    \bm u=
     M_\mathrm{gnd}^{-1}
    \begin{bmatrix}
        f_{xd}\\ \bar f_z\\ \tau_{yd}\\ \tau_{zd}
    \end{bmatrix},
    \label{eq:mixer_ground}
\end{align}
where the $4\times4$ allocation matrix is invertible for the optimized
structure (its upper $1$st, $3$rd, and $4$th rows form $\bm E$, of full row
rank by Proposition~\ref{prop:halfturn}, and the vertical row is independent
for the optimized angles).
When actuator saturation or contact margins become active, the inverse
allocation is replaced by the constrained least-squares allocation
\begin{align}
    \bm u^\ast
    =\argmin_{\bm u\in\mathcal U}~&
    \|\bm E\bm u-\bm w_d\|^2+\rho\|\bm u-\bm u_{\rm nom}\|^2\\
    \text{s.t.}~&m_c^{\rm qs}(\bm u)\ge\underline m,\notag
    \label{eq:allocation_qp}
\end{align}
a small quadratic program that preserves a prescribed contact margin
$\underline m>0$.

%%============================================================================%%
\section{Simulation Validation}

Because the main contribution of this work is the vehicle structure, the simulations evaluate both the hovering capability and the surface-locomotion performance of the proposed configuration. The controllers in Section~V were evaluated using a MuJoCo rigid-body simulation of the two-wheeled drone. The standard untilted configuration and the proposed 4P configuration were tested using identical position- and attitude-control gains.

\begin{figure}[t]
    \centering
    \includegraphics[width=0.9\linewidth]{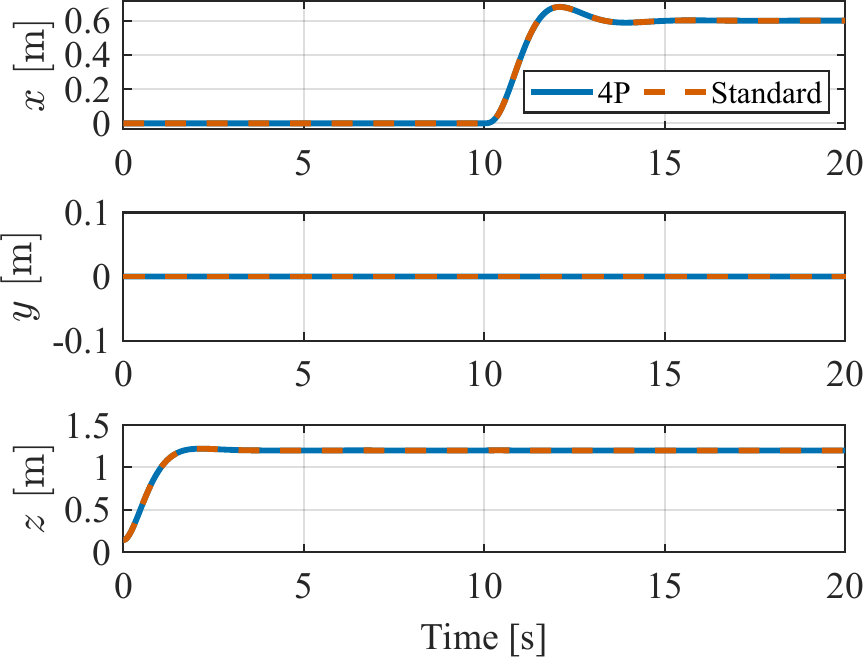}
    \caption{Position responses during the hovering maneuver for the standard and proposed 4P configurations.}
    \label{fig:hover_simulation_position}
\end{figure}

\begin{figure}[t]
    \centering
    \includegraphics[width=0.9\linewidth]{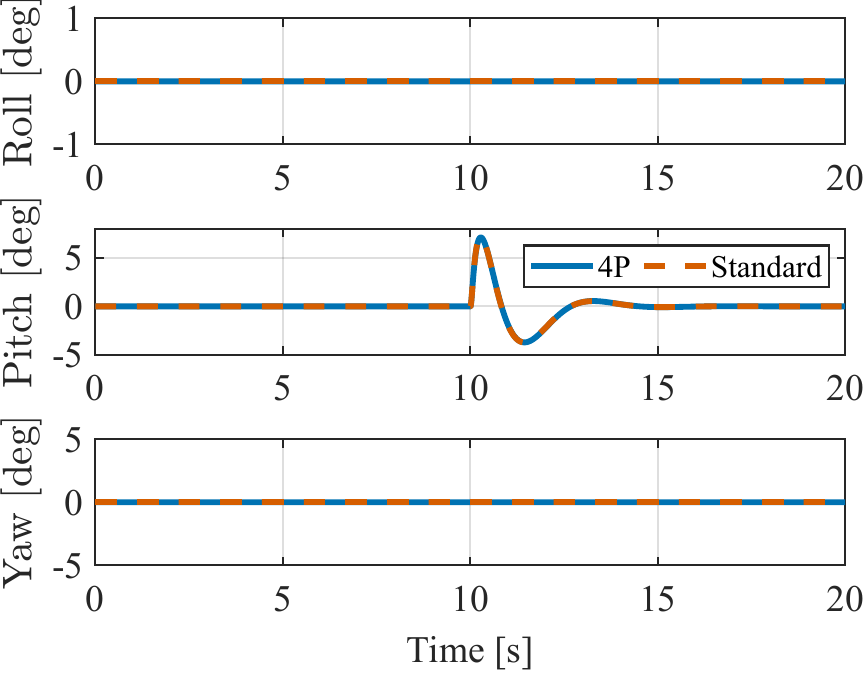}
    \caption{Attitude responses during the hovering maneuver for the standard and proposed 4P configurations.}
    \label{fig:hover_simulation_attitude}
\end{figure}

First, the hovering capability was evaluated in free flight. A target altitude of $z=1.2$~m was given at $t=0$~s, followed by a target position of $x=0.6$~m at $t=10$~s. Figs.~\ref{fig:hover_simulation_position} and \ref{fig:hover_simulation_attitude} show the position and attitude responses, respectively. The responses of the two configurations are nearly identical: both reach the target altitude and subsequently track the horizontal position command while maintaining stable attitude. These results show that the proposed 4P configuration maintains hovering and position-control performance comparable to the standard configuration.

\begin{figure}[t]
    \centering
    \includegraphics[width=0.9\linewidth]{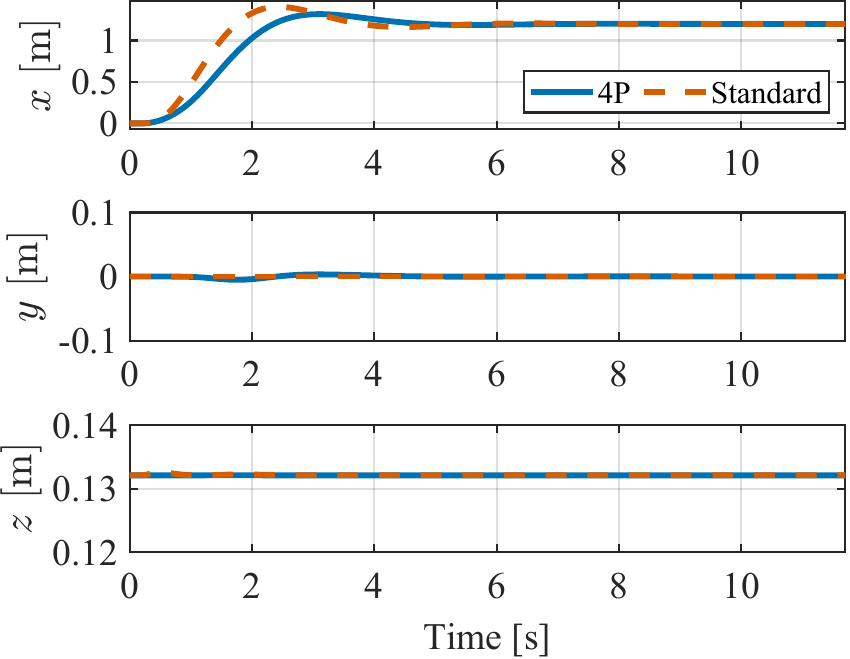}
    \caption{Position responses during the 1.2~m ground-locomotion maneuver for the standard and proposed 4P configurations.}
    \label{fig:ground_simulation_position}
\end{figure}

\begin{figure}[t]
    \centering
    \includegraphics[width=0.9\linewidth]{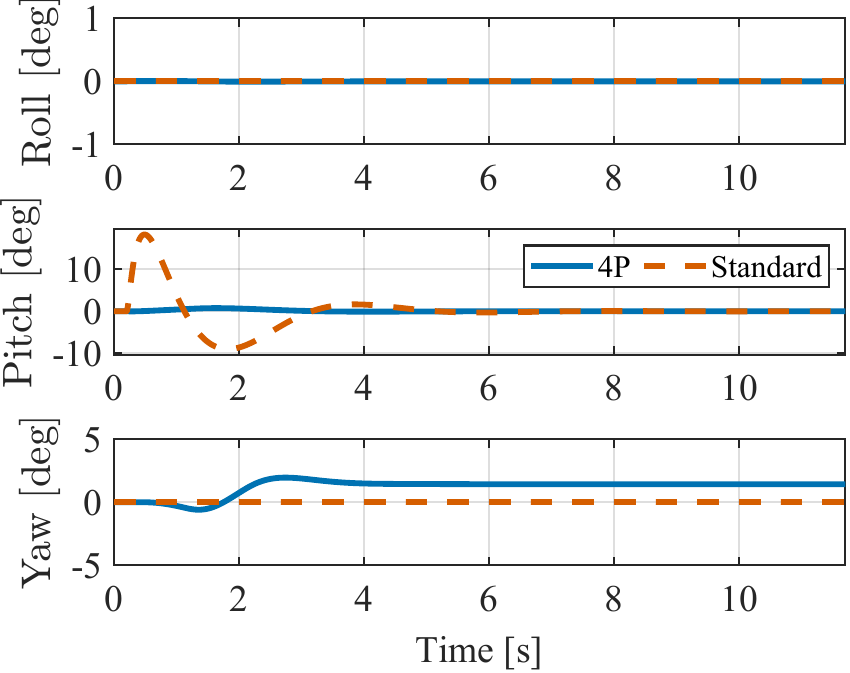}
    \caption{Attitude responses during the 1.2~m ground-locomotion maneuver for the standard and proposed 4P configurations.}
    \label{fig:ground_simulation_attitude}
\end{figure}

Next, the surface-locomotion performance was evaluated from an initially stationary contact state. Both vehicles performed a 1.2~m point-to-point maneuver along the $x$ direction. Figs.~\ref{fig:ground_simulation_position} and \ref{fig:ground_simulation_attitude} compare the position and attitude responses, respectively.
As shown in Fig.~\ref{fig:ground_simulation_position}, both configurations converge to the target position while maintaining their lateral and vertical positions. The proposed 4P configuration reaches a maximum $x$ position of 1.318~m, corresponding to a 9.9\% overshoot, whereas the standard configuration reaches 1.406~m, corresponding to a 17.2\% overshoot. The lateral displacement of the 4P configuration remains within approximately 0.005~m, and the vertical position remains nearly constant after the initial contact transient.

The difference between the two configurations is more apparent in the attitude responses shown in Fig.~\ref{fig:ground_simulation_attitude}. The standard configuration generates tangential acceleration by pitching the body, as predicted by Remark~1, resulting in a pitch excursion from $-8.96^\circ$ to $18.20^\circ$. In contrast, the proposed 4P configuration directly generates tangential force while maintaining a nearly level attitude. Its roll and pitch angles remain within approximately $0.005^\circ$ and $0.75^\circ$, respectively. A small yaw deviation of approximately $2^\circ$ is observed for the 4P configuration during the maneuver. Although some yaw motion remains, the roll and pitch variations are substantially smaller than the pitch variation required by the standard configuration. These results demonstrate that the proposed 4P structure achieves point-to-point surface locomotion without relying on the large body tilting required by the standard configuration, consistent with the reduced-wrench analysis in Section~\ref{sec:hoverability}.

\section{Conclusion}

This paper presented a two-wheeled UAV with tilted rotors for surface locomotion while maintaining a nearly level attitude. Based on reduced-wrench analysis under wheel-contact constraints, a 4P rotor configuration was obtained through structure optimization to provide direct tangential-force authority while retaining hovering capability. 
Simulation results demonstrated both hovering capability and surface locomotion with substantially smaller attitude variation than the standard untilted configuration.

\bibliographystyle{IEEEtran}
\bibliography{reference}

\end{document}